# Exploring Robot Trajectory Planning - A Comparative Analysis of Algorithms and Software Implementations In Dynamic Environments

Arunabh Bora
MS Robotics & Autonomous Systems
University of Lincoln
*Student ID - 27647565*
27647565@students.lincoln.ac.uk

*Abstract*—Trajectory Planning is a crucial word in Modern & Advanced Robotics. It's a way of generating a smooth and feasible path for the robot to follow over time. The process primarily takes several factors to generate the path, such as velocity, acceleration and jerk. The process deals with how the robot can follow a desired motion path in a suitable environment. This trajectory planning is extensively used in Automobile Industrial Robot, Manipulators, and Mobile Robots. Trajectory planning is a fundamental component of motion control systems. To perform tasks like pick and place operations, assembly, welding, painting, path following, and obstacle avoidance. This paper introduces a comparative analysis of trajectory planning algorithms and their key software elements working strategy in complex and dynamic environments. Adaptability and real-time analysis are the most common problems in trajectory planning. The paper primarily focuses on getting a better understanding of these unpredictable environments.

## I. Introduction

Robots are already revolutionizing various industries, automating tasks with advanced technology from basic agriculture to space and medical industries. Within any robotic application, it's essential to establish not just a path but also a motion regulation ensuring the system operates safely and effectively, meeting the task requirements and the robot's limitations [1]. A crucial aspect of robot functionality is Robot Trajectory Planning. It's a fundamental concept in robotics. The word trajectory planning refers to finding the optimal path and motion profile for a robot. Trajectory planning is a subset of motion planning that specifically focuses on generating smooth and feasible trajectories for the robot. Precisely, it involves determining the exact path that the robot's end effector (eg- gripper or working tool ) or joints should follow between each pair of consecutive waypoints along the planned motion path [3].

This paper introduces a novel approach to analyze the differences and gaps between already-developed algorithms and software elements for trajectory planning. The trajectory planning can be computed in both discrete and continuous methods [2][4]. Various planning algorithms are used in Trajectory planning. These include Artificial Potential Field, Sampling-based Planning, Grid-based planning, and Reward-based planning [2].

The problem constraints of trajectory planning can be divided into primarily two major parts – holonomicity and dynamic environments. A holonomic robot has controllable degrees of freedom equal to or greater than its total degree of freedom. With a holonomic robot, it's pretty much easy to control the trajectories, but a non-holonomic robot such as an autonomous car , the compliance of the trajectories is difficult. Similarly, trajectory planning faces more complexities in dynamic environments, where obstacles are in motion. Along with time, the object's movements also change, in this situation calculating a precise trajectory is very challenging.

In this paper, my research question primarily focuses on how the software and algorithms address the challenges of real-time adaptability of trajectory planning in dynamic environments. Specifically, I aim to investigate the latest advancements in trajectory planning algorithms, analyzing their efficacy in dynamic complex environments such as moving obstacles, varying terrain and unpredictable disturbances. By analyzing various researchers' findings, I want to provide some insights into state-of-the-art approaches to achieving agile and responsive trajectory planning in real-world environments.

This paper is divided into many parts, initially the abstract and introduction of the problem statement of the study. The next part is the literature review, it's about all related developments. This part is the most crucial part of the paper, where I am describing all kinds of algorithmic strategies of trajectory planning developments. The next part is the extensive discussion on the literature review. Finally, the paper concludes by summarizing my findings and highlighting potential future directions.

## II. Literature Review

### A. Literature Paper [3]

The paper (Ruscelli,F 2022) introduces Horizon, an open-source framework designed for trajectory optimization in robotics systems. The paper [3] addresses the challenge of development efficient algorithms for autonomous robot routing in urban areas, considering factors like traffic as well as road conditions and pedestrian safety.

The authors present an innovative approach by merging machine learning algorithms with traditional methods for adaptive learning in real-time dynamic environment.

The paper [3] novelty lies in its approach to adaptively learning and optimizing the trajectory in urban environment. And their results are able to demonstrate that with a very simplistic code of instruction that can be translates to a NLP problem, complex motions can be successfully generate by optimization for fast deployment.

One of the major limitations, I have found in the paper is, the approach is relying on historical data and patterns, which may not be always accurate. Because in urban areas, due to the unexpected situations, the algorithm can potentially lead to a suboptimal decision. Moreover, authors also pointing that Horizon currently lacks a well-developed control layer to close the loop on real hardware.

**Table(1) -** *Comparison between state-of-art libraries for trajectory optimization and optimal control [3]*

| Name | Transcription Method | NLP-Solver | Language | Scope |
|---|---|---|---|---|
| Horizon [3] | dms, dc | CasADi solvers, ILQR, GN-SQP | Python | generic |
| DRAKE | dms, dc | IPOPT, custom solvers | C++ | generic |
| Crocoddyl | dms | DDP | C++ | generic |
| OCS2 | dms | DDP, SQP | C++ | generic |
| Control Toolbox | dms | IPOPT, SNOPT, ILQR | C++ | generic |
| FROST [4] | dc | IPOPT, SNOPT, Fmincon2 | MATLAB | gait generation |
| TROPIC | dc | CasADi solvers | C++ | gait generation |
| TOWR | dc | IPOPT, SNOPT | C++ | locomotion |

*B. Literature Paper [4]*

The paper (A. Hereid,2017) [4] addresses the challenge of modeling, optimizing and simulating of robotics system to get a precise trajectory in dynamic environment within the HZD framework.

The paper [4] introduces FROST, an open-source MATLAB toolkit specifically designed for dynamic legged locomotion. FROST is a trajectory planning optimization algorithm. The algorithm can do rapid prototyping of mathematical models and trajectory optimization utilizing feedback controllers and virtual constraint-based motion planning [Fig (1)].

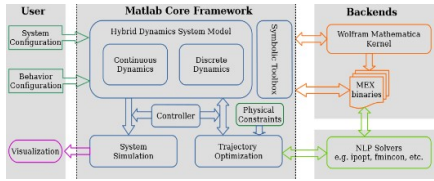

Fig(1)- Block diagram of the FROST algorithm [4] .

The novelty lies in FROST approach is the of modeling hybrid dynamical systems, for which FROST can automatically constructs a multi-phase hybrid trajectory optimization nonlinear programming (NLP) problem. Unlike other methods of trajectory planning frameworks, FROST introduces '*defect variables*' or '*stack variables*' to avoid computing the closed form system dynamics explicitly.

One potential shortcoming of the paper [4] is the results are completely relied on case studies. Although the FROST algorithm convergence faster than others, but the paper does not delve deeply into the potential limitations in terms of scalability.

*C. Literature Paper [5]*

The paper (Chen L, 2022) [5] proposes an approach to optimize trajectory planning using Deep Reinforcement Learning in uncertain environment.

The traditional methods of Trajectory Planning struggle with complex and high-dimensional environments with uncertainties. However, this approach can solve these issues with a significant computation design. The paper [5] is using two DRL algorithms – Software-Actor-Critic (SAC) and Deep Deterministic Policy Gradient (DDPG) for robot training. This reinforcement learning methodology allow the robot to get a better idea about environment.

The novelty of the paper [5] relies on the comparison study between two algorithms. The study summarizes that SAC achieves faster learning compare to DDPG in complex simulation environment.

The paper [5] acknowledge some limitations, that evaluation is done in entirely in a simulated environment. So, for real environment, there maybe need some extra methodological approach to optimize the trajectory.

*D. Literature Paper [6]*

The paper (Dai, C, 2020) [6] addresses the challenge of smooth Trajectory planning. The goal is the minimize the jerk (rate of change of acceleration) of Robot trajectory.

The primary contributions of the paper [6] include a local filter for jerk minimization while considering other hardware, a greedy algorithm to be applied to a path with many waypoints and an adaptive sampling strategy for effectively learning a collision-indication for the robotic trajectory fabrication.

The novelty of the paper [6] lies in the use of a combined approach for trajectory planning. The paper [6] is able to achieve a really efficient methodology for trajectory planning by utilizing these algorithms- Adaptive Greedy Algorithm (AGA) & Adaptive Sampling Strategy (ASA). Where AGA iteratively optimizes the jerk along the path, focusing on regions with high initial jerk values, on the other hand ASA uses support vector machine to formulate a collision indicating function, and it learns from that function to avoid the real-time computational complexities.

The paper [6] acknowledges the limitations of the algorithm also. The algorithm sometimes fails to handle the noisy oriented data for waypoints. Jerk constraints might not be met under significant noise. And most importantly, the paper focuses primarily on simulation not explore the real-world implementation.

*E. Literature Paper [7]*

The paper (Kalakrishnan, M,2011) [7] address the challenge of trajectorial motion planning for robots in complex and dynamic environment. The paper seek a solution for smooth trajectorial path considering constraints and minimum energy requirements.

The primary contribution of the paper is the STOMP algorithm (Stochastic Trajectory Optimization for Motion Planning), which use a series of noisy trajectories to explore the space around of initial trajectory, then employing a cost function that combines the all factors, the algorithm iteratively selects the best trajectory path.

The novelty of the paper [7] is the core idea of using noisy trajectories for exploration and optimization trajectory using the developed STOMP algorithm. The paper [7] demonstrates that the STOMP works at both in simulation and mobile manipulator.

One of the major limitations is that the author suggests that this algorithm needs minor parameters finetuning, but this can't be always true in high-dimensional complex environments. Real-world noise and sensor system delays are not extensively discussed in the paper.

*F. Literature Paper [8]*

The paper (Zhao, 2020) [8] addresses the challenge of ensuring safe navigation for autonomous vehicle. The paper [8] primarily focuses on a two-layered control system for real-time re-planning the trajectorial motion path in dynamic environment with obstacles.

The primary contributions of the paper [8] are the double-layer controller architecture for active collision avoidance and the 2-DOF vehicle dynamics model.

The novelty of the paper is the innovative approach of two layer controller architecture, where the upper layer employs the anti-collision trajectory re-planer based MPC algorithm and the lower layer employs a trajectory tracking controller using fuzzy adaptive PID method.

Although, the authors of the paper [8] evaluates their algorithm in real-time basis, but still there are some limitations like it is applied on only 2-DOF vehicle model, which might not be capture the full dynamics of the vehicle in complex maneuvers. Another limitation is the paper doesn't explore multiple obstacles.

*G. Literature Paper [9]*

The paper (Mir, I, 2022) [9] identifies some traditional and progressive trajectory planning methods for autonomous vehicles.

The primary contribution of the paper [9] is the comparative analysis of already existing trajectory planning algorithms methods and optimization techniques for autonomous vehicles. Also, the paper identifies the future directions of trajectory planning research, including multi-robot trajectory planning methodology.

The novelty of the paper [9] can be expressed as the authors compare all possible methodologies in all possible dynamic environments (aerial, ground, underwater) unlike other research comparisons. The authors also highlight the importance of new optimization techniques like Aquila Optimizer.

The paper [9] focuses only on a broader perspective of every algorithm, it's not discussing the real-time complexities of the algorithm. Additionally, the paper [9] does not directly compare the performance of the same algorithm in different environments.

*H. Literature Paper [10]*

The paper (Q. Zhang, 2017) [10] proposes an innovative approach to generate a smooth and minimal running time robot trajectory planning.

The primary contribution of the paper [10] is the Genetic Chaos Optimization Algorithm, which leverages both the genetic algorithm (good global search) and the chaos algorithm (effective local search).

The novelty of the paper [10] lies in the innovative algorithm, where initially a fifth-degree polynomial curve is employed to accommodate the starting point, end point and intermediate points of the identified path, then using the algorithm, a smooth and optimized trajectory plan is generated to implement at the robot end effector.

I have found the major limitation is that the paper [10] doesn't provide a direct comparison between the existing methods and their proposed method in terms of speed and computational efficiency.

*I. Literature Paper [11]*

The paper (X. Zhang, 2019) [11] introduces an approach for safe trajectory planning for UAVs in complex 3D complex environments with multiple objects.

The primary contribution of the paper [11] is the combine algorithm which integrates a global path planner (HLT*) and a nonlinear model predictive control(MPC) in 3D trajectory planning.

The novelty of the paper [11] lies in the innovative approach for the integrated algorithm, which initially compute the waypoints using HLT* then by using NMPC and sequential quadratic programming it achieves the smooth and collision free trajectories while satisfying all constraints.

The algorithm might need more suitable solution for real-time 3D environment. The paper [11] doesn't discuss about real-time solution. Additionally, the simulation is validated within UAVs only.

*J. Literature Paper [12]*

The problem statement of the paper (P. Savsani, 2016) [12] is to conduct a study of seven different metaheuristic trajectory algorithms for a three-revolute (3R) robotic arm.

The primary contributions of the paper [12] are the evaluation of metaheuristic algorithms and identifying the optimal solutions in free and obstacle workspaces.

The novelty of the paper [12] lies in application of several algorithms for optimization study relevance of real-time world scenarios. The algorithms are ABC algorithm, BBO, GSA, CS, FA, BA, and TLBO, where the TLBO, ABC and CS algorithms outperform others in term of best solution, convergence and statistical significance.

The primary limitation of the paper [12] is that it doesn't discuss how these algorithm might perform in other robots also beyond 3R arm. Although the paper discusses some real-time scenarios, still crucial results are shown in the simulation frame.

### III. DISCUSSIONS

The below table [Table(1)] is the overall summary of the literature papers. From the comparisons, a striking observation is the prevalent focus on the simulated environments. While these studies [3,4,7, 10] offer a valuable contribution to algorithm development, the gap between simulation and real-world implements remains a significant challenge. Factors like sensor noise, hardware limitations, and unpredictable dynamic changes are not often adequately addressed. Despite the limitations, the literature review also highlights promising directions for real-time adaptability. The incorporation of deep reinforcement learning [5] holds a significant promise for enabling the real-time environment. Hybrid approaches such as the paper [8] offer a robust approach to safe and adaptable implementations. Exploring new algorithms like 'Aquila Optimizer' (introduced in [9]), could potentially improve the efficiency of real-time planning. The computational efficiency also plays a crucial role in the trajectory planning for real-time planning [10]. So, future research should prioritize the development of computationally efficient algorithms. Moreover, developing algorithms which can balance the trade-off between optimality and speed could be a better approach for future directions to achieve a solution within a limited timeframe.

Multi-robot trajectory planning (brought up in [9]) could be a potential research direction for the future.

**Table (1) – Overall Comparison of Literature Reviews**

| Paper | Challenge Addressed | Proposed Approach | Novelty | Limitations |
|---|---|---|---|---|
| [3] | Trajectory optimization for robots in urban environments | Horizon framework, combining machine learning and traditional methods | Learning and optimizing trajectory in real-time using NLP | Relies on historical data, lacks well-developed control layer |
| [4] | Precise trajectory planning in dynamic environments | FROST algorithm for trajectory optimization | Automatic construction of multi-phase NLP | Limited results on scalability, case study based |
| [5] | Trajectory planning in uncertain environments | Deep Reinforcement Learning (DRL) for robot training | Comparison of SAC and DDPG algorithms for faster learning | Evaluation is done entirely in simulation |
| [6] | Smooth trajectory planning with minimal jerk | Combined approach using (AGA) and (ASA) | Efficient jerk minimization and real-time collision avoidance | Fails with noisy data, limited to simulation |
| [7] | Smooth trajectory planning with minimal energy in complex environments | STOMP (Stochastic Trajectory Optimization for Motion Planning) algorithm using noisy trajectories | Exploration and optimization using noisy trajectories | Requires parameter tuning, limited discussion on real-world noise |
| [8] | Safe navigation for autonomous vehicles in dynamic environments | Double-layer control system with MPC based replanning | Innovative two-layer architecture for real-time replanning | 2-DOF vehicle model limits dynamics, not suitable for multiple obstacles |
| [9] | Comparative analysis of trajectory planning methods for autonomous vehicles | Analysis of existing methods and future directions | Compares methods across various environments (aerial, ground, underwater) | Focuses on broad perspective, lacks real-time complexity analysis |
| [10] | Smooth and minimal running time robot trajectory planning | Genetic Chaos Optimization Algorithm | Innovative algorithm for smooth, optimized trajectory generation | No direct comparison with existing methods in terms of efficiency |
| [11] | Safe trajectory planning for UAVs in 3D environments | Combined approach using HLT* global planner and NMPC | Integration of global path planning and real-time control | Needs efficient solution for real-time 3D environments |
| [12] | Trajectory planning for 3-revolute robotic arm | Evaluation of metaheuristic algorithms for optimal solutions | Analysis of ABC, BBO, GSA, CS, FA, BA, and TLBO algorithms | Limited to 3R arm, simulation focused with limited real-world validation |

## IV. CONCLUSION

As technology progresses, the landscape of trajectory planning in advanced and intelligent robotics systems is shifting towards increasingly hybrid and adaptive solutions. The incorporation of artificial intelligence and machine learning methods are making possible for enhancing the trajectory planning algorithms.

In this paper I discuss about several existing methodologies and potential future directive technologies for robot trajectory planning in dynamic or high dimensional complex environment. While I aim to gain insights into the algorithms' strategy and applications, I also highlight the limitations or shortcomings of the existing technologies. I have briefly explained the approaches and strategic points of several algorithms, including Horizon, FROST, STOMP, GCOA and many more. The primary limitation of my survey is that it majorly covers the trajectory planning methods in simulative environmental studies. In robotics, real-time dynamics studies along with hardware configuration are also crucial for gaining deeper knowledge. Based on my research question, I can conclude that integration of traditional and advanced machine learning technology can significantly impact this field.